\documentclass[letterpaper, 10 pt, conference]{ieeeconf}  
\let\labelindent\relax

\IEEEoverridecommandlockouts                              

\usepackage{microtype}
\usepackage{cite}
\usepackage{amsmath,amssymb,amsfonts}
\usepackage{algorithmic}
\usepackage{graphicx}
\usepackage{textcomp}
\usepackage{xcolor}
\usepackage{hyperref}
\usepackage{tabularx} 
\usepackage{booktabs} 
\usepackage{multirow}
\usepackage{etoolbox}      
\def\BibTeX{{\rm B\kern-.05em{\sc i\kern-.025em b}\kern-.08em
    T\kern-.1667em\lower.7ex\hbox{E}\kern-.125emX}}
\usepackage{cite} 
\usepackage{enumitem}
\usepackage{float}

\title{\LARGE \bf
FASR-Net: Unsupervised Shadow Removal Leveraging Inherent Frequency Priors
}

\author{Tao Lin$^{1}$, Qingwang Wang$^{2,*}$, Qiwei Liang$^{3}$, Minghua Tang$^{4}$, and Yuxuan Sun$^{5}$%
\thanks{This work was supported by Yunnan Fundamental Research Projects}
\thanks{$^{1}$Tao Lin, Qingwang Wang, Minghua Tang, Yuxuan Sun are with Faculty of Information Engineering and Automation, Kunming University of Science and Technology, Kunming, China}
\thanks{$^{3}$Qiwei Liang is with College of Mechatronics and Control Engineering, Shenzhen University, Shenzhen, China}
\thanks{$^{*}$Corresponding author: Qingwang Wang. E-mail: wangqingwang@kust.edu.cn}
}

\begin{document}

\maketitle
\thispagestyle{empty}
\pagestyle{empty}

\begin{abstract}
Shadow removal is challenging due to the complex interaction of geometry, lighting, and environmental factors. Existing unsupervised methods often overlook shadow-specific priors, leading to incomplete shadow recovery. To address this issue, we propose a novel unsupervised Frequency Aware Shadow Removal Network (FASR-Net), which leverages the inherent frequency characteristics of shadow regions. Specifically, the proposed Wavelet Attention Downsampling Module (WADM) integrates wavelet-based image decomposition and deformable attention, effectively breaking down the image into frequency components to enhance shadow details within specific frequency bands. We also introduce several new loss functions for precise shadow-free image reproduction: a frequency loss to capture image component details, a brightness-chromaticity loss that references the chromaticity of shadow-free regions, and an alignment loss to ensure smooth transitions between shadowed and shadow-free regions. Experimental results on the AISTD and SRD datasets demonstrate that our method achieves superior shadow removal performance.
\end{abstract}

\section{Introduction}
\label{sec:intro}
Shadows arise from light obstruction and can obscure visual information in images~\cite{guo2024single}. This presents challenges in autonomous driving, virtual reality, and augmented reality. For autonomous vehicles, shadows can conceal road markings or objects, causing errors and compromising safety. In virtual and augmented reality environments, shadows disrupt immersion and interaction by altering object shapes.
\begin{figure}[t] 
    \centering
    \includegraphics[width=\columnwidth]{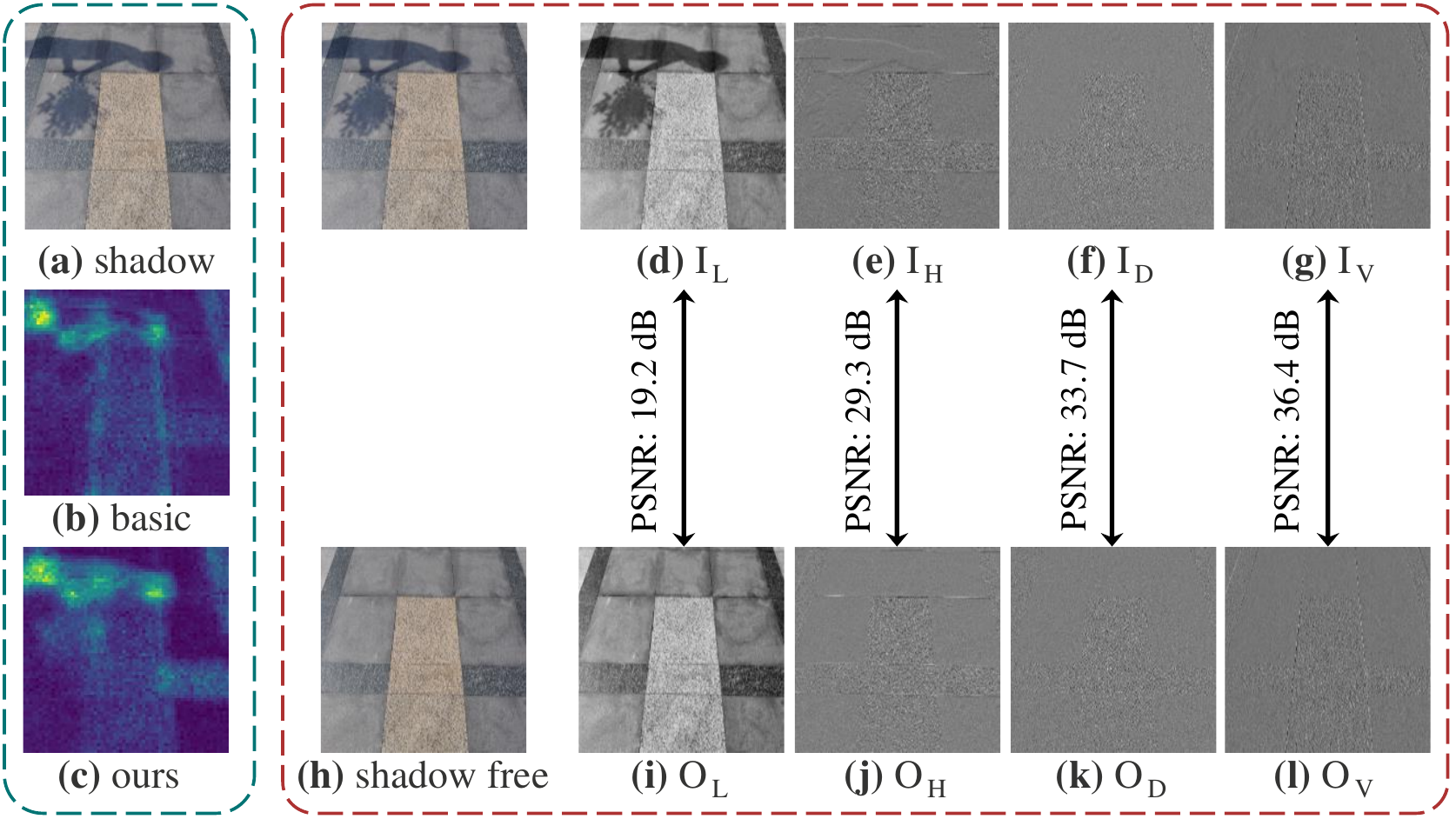} 
    \vspace{-1mm}
    \caption{\textbf{Module heatmap comparison and PSNR analysis.} The left (green) part shows the heatmap comparison of the basic generator framework after the Wavelet Attention Downsampling Module (WADM) is added ((b) and (c)). The right (red) section presents the wavelet transform of shadow and shadow-free images, highlighting low-frequency ((d) and (i)) and high-frequency components ((e-g) and (j-l)). PSNR shows that the D and V components have the highest similarity.}
    \label{fig:myfigure}
\end{figure}
\noindent 
Given the challenges of shadows in applications, deep learning~\cite{liu2021g2rshadow,liu2023decoupled,hu2018direction,duan2022shadow,koutsiou2024sushe,dong2024shadowrefiner} has achieved great success in shadow removal. Using large amounts of data, it can learn the complex relationships between shadowed and shadow-free areas well and improve image visual quality impressively. However, gathering a large amount of high-quality training data is difficult in reality. The varying sunlight and sky in different scenes make collecting shadowed and shadow-free image pairs, essential for most supervised deep-learning methods, extremely costly. Some studies adopt synthetic datasets to tackle data scarcity. Despite color and pixel alignment, they struggle to capture real-world subtleties like lighting changes and complex textures. Thus, models trained on these datasets often have unsatisfactory results and weak adaptability to real-world situations. This challenge is further exacerbated by the limitations of existing supervised methods, which struggle to address diverse soft shadows and complex scene structures. 
\begin{figure*}[!t] 
    \centering
    \includegraphics[width=\textwidth]{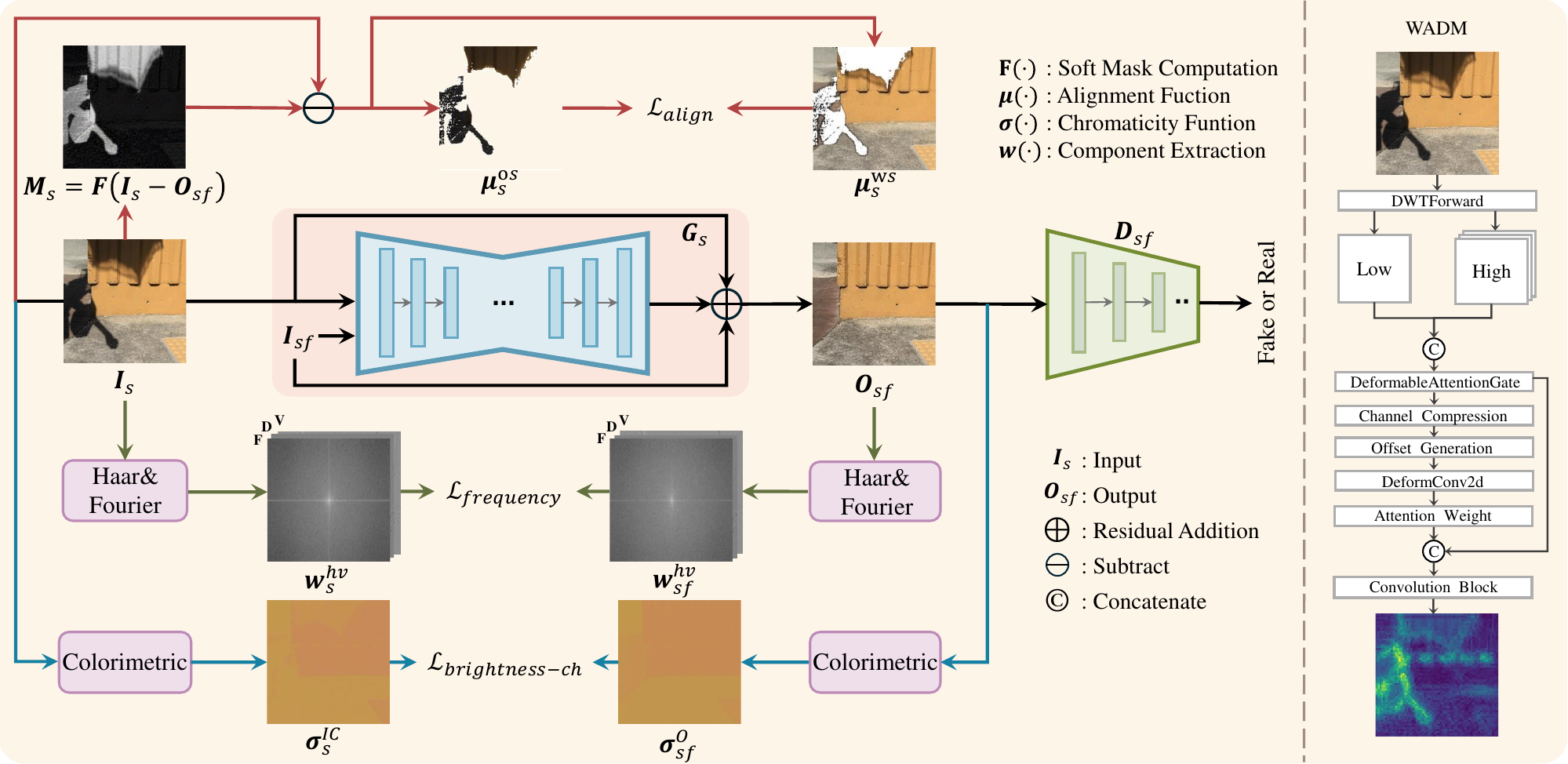}
    \setlength{\abovecaptionskip}{0pt} 
    \setlength{\belowcaptionskip}{0pt} 
    \caption{\textbf{Overall pipeline of our network.} A network shadow and shadow-free domains use a shadow removal generator \( G_s \) to transform \( I_s \) to \( O_{\text{sf}} \) and reconstruct \( I_{\text{sf}} \). It features a WADM in the downsampling stage and proposes new losses like \( \mathcal{L}_{\text{brightness-ch}} \), \( \mathcal{L}_{\text{frequency}} \), and \( \mathcal{L}_{\text{Align}} \) for enhanced shadow removal.}
    \label{fig:yfigure}
\end{figure*}
Unsupervised deep learning methods have emerged as promising alternatives to address the limitations of traditional shadow removal techniques. Among these, GAN-based methods~\cite{zeng2024semantic} have gained significant popularity due to their ability to generate highly realistic images without requiring extensive labeled datasets. Notably, Mask-ShadowGAN~\cite{hu2019mask} by Hu \textit{et al.} pioneered unsupervised shadow removal using adversarial training. However, this method struggles to completely eliminate shadows and effectively handle soft shadows, as cycle consistency alone does not ensure robustness across diverse scenarios. Additionally, Vasluianu \textit{et al.}~\cite{vasluianu2021shadow} addressed dataset inaccuracies by employing blurring for improved color consistency, relying solely on perceptual losses. Building upon these works, DC-ShadowNet~\cite{jin2021dc} combines the strengths of Mask-ShadowGAN, introducing a domain classifier that adeptly manages both soft and hard shadows. Despite these advancements, existing methods primarily depend on network structures and constraints, overlooking the inherent characteristics and complexities of shadows themselves. 

Shadows possess distinct frequency-domain characteristics: high-frequency components capture edges and details, while low-frequency components represent the overall shape and area~\cite{yu2023fsr}. Ignoring these characteristics can lead to the loss of edge details or difficulties in distinguishing shadows, ultimately affecting the quality of shadow removal. Leveraging insights, we can maintain essential image information while reducing the spatial resolution of feature maps. This approach enhances the accurate identification of shadow regions, even without a mask, thus improving shadow removal in complex scenes (see Fig. 1). Moreover, as demonstrated by wavelet transform decomposition, the similarity in high-frequency components effectively guides the shadow removal process.

Inspired by integrating frequency-domain knowledge into unsupervised learning, we propose a novel Frequency Aware Shadow Removal Network (FASR-Net) for unpaired shadow-free input images. Firstly, we incorporate the Wavelet Attention downsampling Module (WADM) during the downsampling phase of the generator network. This module takes advantage of frequency prior derived from the Haar wavelet~\cite{xu2023haar}. With the augmentation of cascade pooling and the calculation of convolution offsets, the capacity of the generator to capture intricate details in images is significantly boosted. After applying wavelet transform decomposition to the input image pairs, we discover specific similarities among the diagonal, vertical components, and the spectrum. These similarities prompt the design of a shadow frequency loss, which is crucial for restoring frequency features and enhancing shadow removal. Meanwhile, unlike the shadow-free chromaticity loss~\cite{jin2021dc}, our shadow brightness-chromaticity loss uses the brightness space as a guide to better preserve color brightness in the shadow-free state. A soft shadow mask can also be obtained by subtracting the shadow-free image from the input shadow image. The shadow mask alignment loss can be used to adjust the shadow and shadow-free regions, achieving alignment of the related image distributions. In the light of above, our contributions can be outlined as follows:

\begin{itemize}
  \item We propose the Wavelet Attention Down Module (WADM) for the generator downsampling stage, leveraging Haar wavelet frequency priors, cascade pooling, and convolution offset calculations to enhance the model's ability to capture intricate image details.
  \item We develop several novel loss functions: frequency loss for enhanced accuracy using high-frequency details, brightness-chromaticity loss for guiding removal through chromaticity, and mask alignment loss for preserving details by aligning shadowed and shadow-free regions.
\end{itemize}

\def\BibTeX{{\rm B\kern-.05em{\sc i\kern-.025em b}\kern-.08em
    T\kern-.1667em\lower.7ex\hbox{E}\kern-.125emX}}
\setlength{\textfloatsep}{10pt plus 1.0pt minus 2.0pt}
\setlength{\floatsep}{10pt plus 1.0pt minus 2.0pt}
\let\OLDthebibliography\thebibliography
\renewcommand\thebibliography[1]{
  \OLDthebibliography{#1}
  \setlength{\parskip}{0pt}
  \setlength{\itemsep}{0pt plus 0.3ex}
}
\section{METHODOLOGY}
\label{sec:METHODOLOGY}
Refer to Fig. 2 for an illustration of our method. The network has two domains: shadow \( I_s \) and shadow-free \( I_{sf} \), with a shadow removal generator \( G_s \) that transforms \( I_s \) into \( O_{sf} \) and reconstructs \( I_{sf} \). In the generator and discriminator \( D_s \) framework containing the domain classifier, we propose a Wavelet Attention Down Module (WADM), which uses Haar wavelets to introduce frequency information and combines maximum pooling, average pooling, and deformable attention to enhance shadow detail capture. To guide $G_s$ for shadow removal, we further propose new losses. The brightness-chrominance loss $\mathcal{L}_{\text{brightness-ch}}$ operates in the LAB color space. Guided by $\sigma_s^{IC}$ in the brightness space of $I_s$, we reduce the shadows of the L/B channels by mean filtering, then employ principal component analysis and entropy minimization to generate a shadow-free chromaticity map in the logarithmic chromaticity space. The shadow frequency loss $\mathcal{L}_{\text{frequency}}$ is guided by the high-frequency components $D$ and $V$ in the wavelet transform and $F$ in the Fourier transform. Lastly, the shadow mask alignment loss $\mathcal{L}_{\text{Align}}$ adjusts the statistical characteristics of the masked area in the generated image to match the unmasked area of the target image.

\subsection{Wavelet Attention Downsampling Module}
\textbf{Haar Wavelet Transform:} When applying the Haar wavelet transform~\cite{porwik2004haar} to a 2D image of resolution \(H \times W\), the image is treated as a 2D signal. The process performs 1D Haar transforms on each row and column. Low-pass filter \(H\) and high-pass filter \(H_1\) reduce data length from \(L\) to \(\frac{L}{2}\), extracting low and high-frequency information. This yields four \(\frac{H}{2} \times \frac{W}{2}\) components: approximate component \(A\) and detail components \(H\), \(V\), and \(D\) in horizontal, vertical, and diagonal directions. This lossless transform encodes information into the channel dimension, increasing channels from \(C\) to \(4C\).

\textbf{Deformable Method And Z-Pool:} In deformable attention mechanisms~\cite{xia2022vision}, the input feature map \([B, C, H, W]\) undergoes Z-Pool~\cite{misra2021rotate}, which applies max-pooling and average-pooling along the channel dimension. The max-pooling is given by \( z_{\text{max}}[b, c, i, j] = \max_{k \in R_{d}} z[b, k, i, j] \), and the average-pooling by \( z_{\text{avg}}[b, c, i, j] = \frac{1}{|R_{ij}|} \sum_{k \in R_{d}} x[b, k, i, j] \), where \( R_{ij} \) is the neighborhood of \((i, j)\) and \( b \) is the sample index. These pooled features are concatenated, producing a feature map \([B, 2C, H, W]\), reducing channels and fusing statistical data. Next, compress calculates the offset through the convolutional layer. Its calculation formula is:
\begin{equation}
\Delta[b, c, i, j] = \sum_{m=0}^{M-1} \sum_{n=0}^{N-1} w_{mn} \cdot z[b, c, i+m, j+n],
\end{equation}
where $\Delta$ is the offset, $w_{mn}$ are convolutional weights, and $M, N$ are kernel dimensions. This offset and the compressed feature are input to a deformable convolutional layer. The output of the deformable convolution is normalized by $\sigma(z_{\text{out}}) = \frac{1}{1+e^{-z_{\text{out}}}}$ to obtain the attention weight scale. Finally, the original feature map $z$ is multiplied element-by-element with scale to achieve attention weighting.

\subsection{Shadow Frequency Loss}
The shadow frequency loss combines the advantages of the horizontal-vertical and diagonal component loss \(\mathcal{L}_{VD}\) and the focal frequency loss~\cite{jiang2021focal} \(\mathcal{L}_{FF}\).

\(\mathcal{L}_{VD}\) is based on the comparison of wavelet-transformed shadowed and shadow-free images. Haar wavelets extract vertical ($V_f$,$V_r$) and diagonal ($D_f$,$D_r$) high-frequency components. The loss for these components with the formula:
\begin{equation}
\mathcal{L}_{VD} = \frac{1}{n}\sum_{i=1}^{n} (cV_f(i) - cV_r(i))^2 + \frac{1}{n} \sum_{i=1}^{n} (cD_f(i) - cD_r(i))^2,
\end{equation}
where \( n \) is the sample count, and \( c \) is a calculation constant. Minimizing \( \mathcal{L}_{VD} \) aligns high-frequency components of generated and real images, which is important for forming the complete shadow frequency loss function.

Besides, we use the 2D-DFT to decompose the image $f(x,y)$ into orthogonal sine and cosine functions to enhance and optimize different frequency components in the image frequency domain. The spectral coordinates $(u,v)$ determine the angular frequency related to spatial frequency. Considering amplitude and phase information, we map frequency values to vectors to calculate the frequency distance between the real image $F_r(u,v)$ and the generated image $F_f(u,v)$.

Directly using the frequency distance as a loss function can not handle hard-to-synthesize frequencies, we introduce a spectral weight matrix $w(u,v)=|F_r(u,v)-F_f(u,v)|^\alpha$ ($\alpha = 1$). This adjusts frequency loss distribution and reduces weights of easily synthesized frequencies. The Hadamard~\cite{horadam2012hadamard} product of the spectral weight matrix and the frequency distance matrix to obtain the focal frequency loss:
\begin{equation}
\mathcal{L}_{FF}=\frac{1}{MN}\sum_{u = 0}^{M - 1}\sum_{v = 0}^{N - 1}w(u,v)|F_r(u,v)-F_f(u,v)|^2.
\end{equation}

We combine \( \mathcal{L}_{VD} \) and \( \mathcal{L}_{FF} \) to form the shadow frequency loss function \( \mathcal{L}_{frequency} \) with coefficients \( \lambda_1 \) and \( \lambda_2 \):
\begin{equation}
\mathcal{L}_{frequency} = \lambda_1 \mathcal{L}_{VD} + \lambda_2 \mathcal{L}_{FF}.
\end{equation}

\subsection{Shadow Brightness-Chromaticity Loss}
The shadow-free chromaticity loss~\cite{jin2021dc} becomes less effective at creating a shadow-free map when angular hard shadows from a point light source blend with the background. We introduce a brightness-chromaticity loss to enhance shadow region identification by utilizing the LAB space to separate color and luminance components. In the LAB space, shadows appear darker in the L channel and more prominent in the B channel due to their absorption of warm light.
\begin{figure}[t] 
    \centering
    \includegraphics[width=\columnwidth]{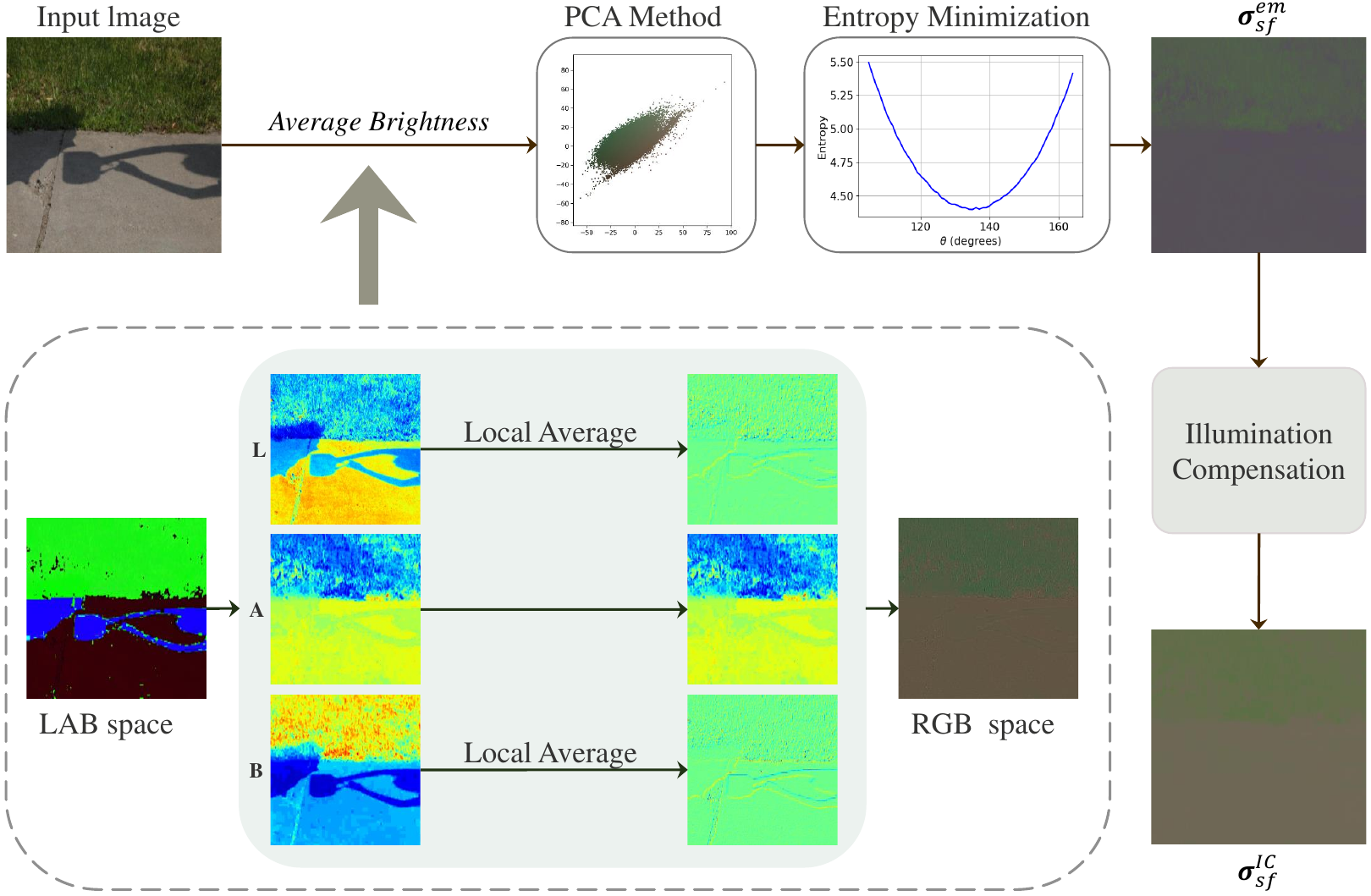} 
    \caption{\textbf{Average brightness and illumination compensation.} The lower left section is the average brightness pipeline where we process the L/B channels of the image in LAB space. After applying PCA and minimizing the entropy, we obtain \( \sigma_{sf}^{em} \). Besides, illumination compensation is performed on the image to obtain \( \sigma_{sf}^{IC} \) that is closer to the color brightness of the input image.}
    \label{fig:myfigure5}
\end{figure}

To reduce the shadow effect in the L/B channels, we first calculate a shadow-free perceived image by using a 3x3 mean filter to average the brightness of the shadowed image~\cite{chen2021canet}. The specific calculation is as follows: for a pixel \((m,n)\), its shadow-free image is given by \( I_{m,n} = I_{m,n} - P_{\text{mean}} + I_{\text{avg}} \), where \( P_{\text{mean}} \) represents the mean brightness of a \(3 \times 3\) patch around the pixel \((m,n)\), calculated by \( P_{\text{mean}} = \frac{1}{N} \sum_{(m,n) \in P} I_{m,n} \). Here, \( P \) is the \(3 \times 3\) patch around the pixel \((m,n)\), \( N \) is the total number of pixels in the patch, \( I_{m,n} \) is the brightness value of the pixel \((m,n)\), and \( I_{\text{avg}} \) is the average brightness value of the entire image.

After processing the luminance of the L/B channels, we revert the image to the RGB space. Given potential color discrepancies between shaded and non-shadowed areas, the PCA method \cite{mackiewicz1993principal} is adapted to compress the color dimensions from 3D to 2D. The shadow-free perceived image is fed into the log chromaticity space following \cite{finlayson2009entropy}. The projection direction is ascertained by calculating and minimizing entropy, yielding a shadow-free colorimetric map. An illumination compensation approach is employed to rectify color bias during projection. Pixels corresponding to the shadow-free region of the input image are sampled from the pre-generated shadow-free perceived image, facilitating the creation of a more precise shadow-free colorimetric map and enhancing the overall quality of shadow removal. As shown in Fig. 4, its color brightness is more similar to the input image than the shadow-free chromaticity loss.\looseness=-1

\begin{figure}[t] 
    \centering
    \setlength{\tabcolsep}{1pt} 
    \renewcommand{\arraystretch}{0} 

    \begin{tabular}{cccccccc}
        \includegraphics[width=0.093\textwidth]{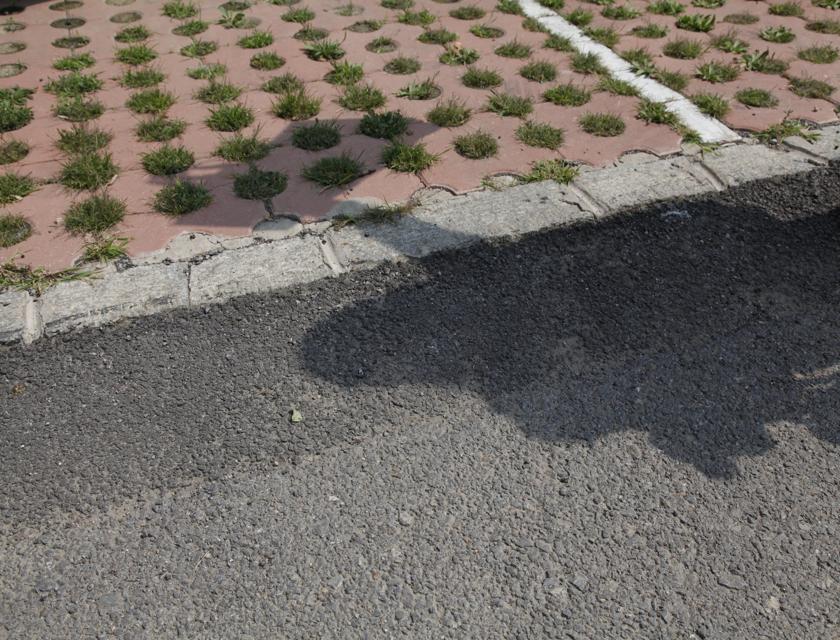} &
        \includegraphics[width=0.093\textwidth]{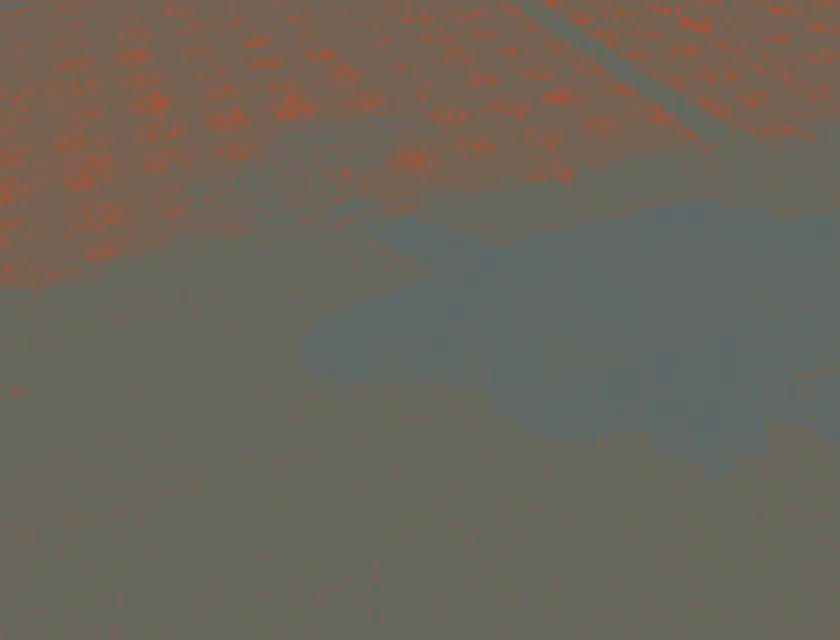} &
        \includegraphics[width=0.093\textwidth]{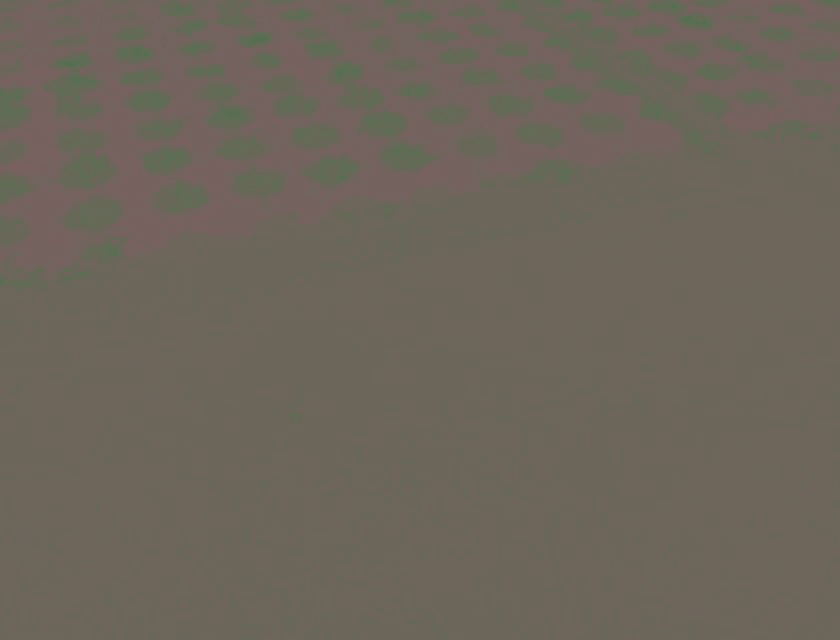} &
        \includegraphics[width=0.093\textwidth]{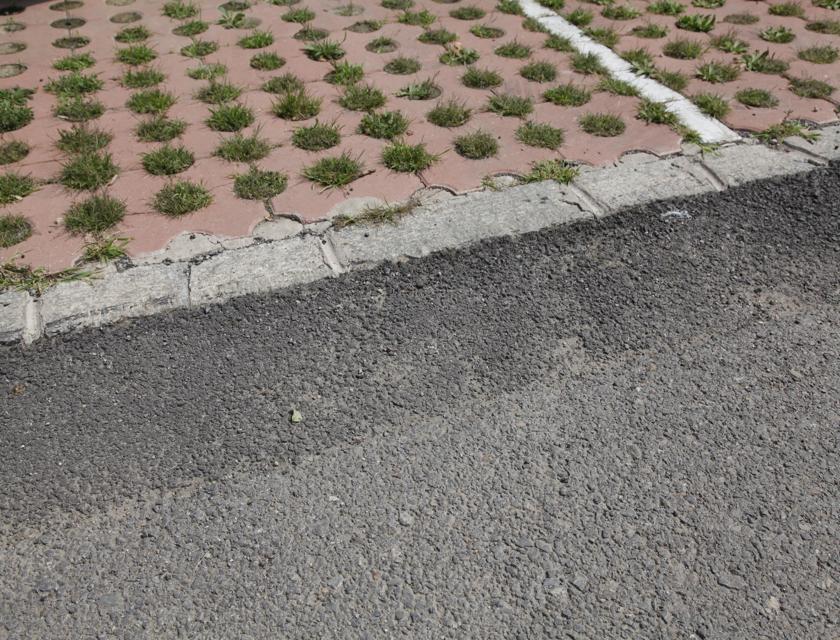} &
        \includegraphics[width=0.093\textwidth]{4/3.png} &
    \end{tabular}\\[0.7ex]

    \begin{tabular}{cccccccc}
        \includegraphics[width=0.093\textwidth]{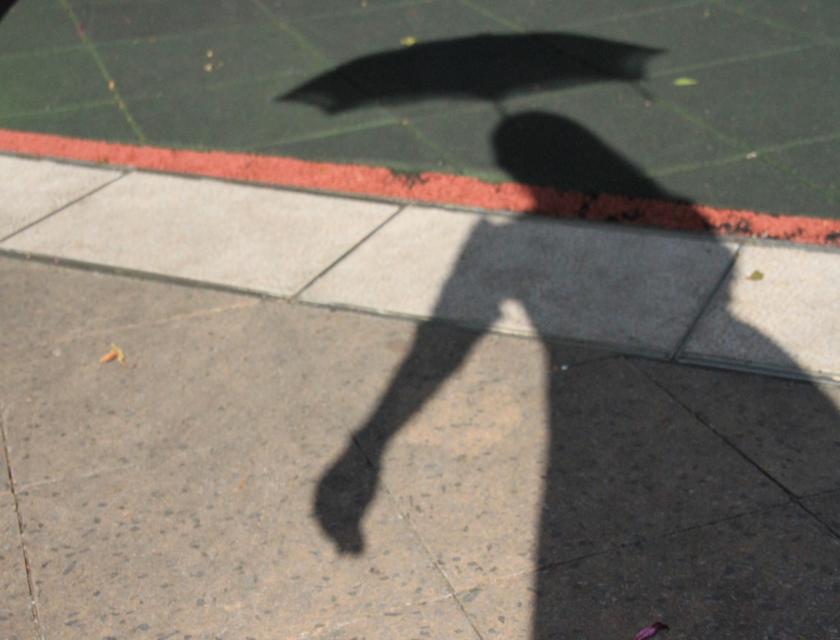} &
        \includegraphics[width=0.093\textwidth]{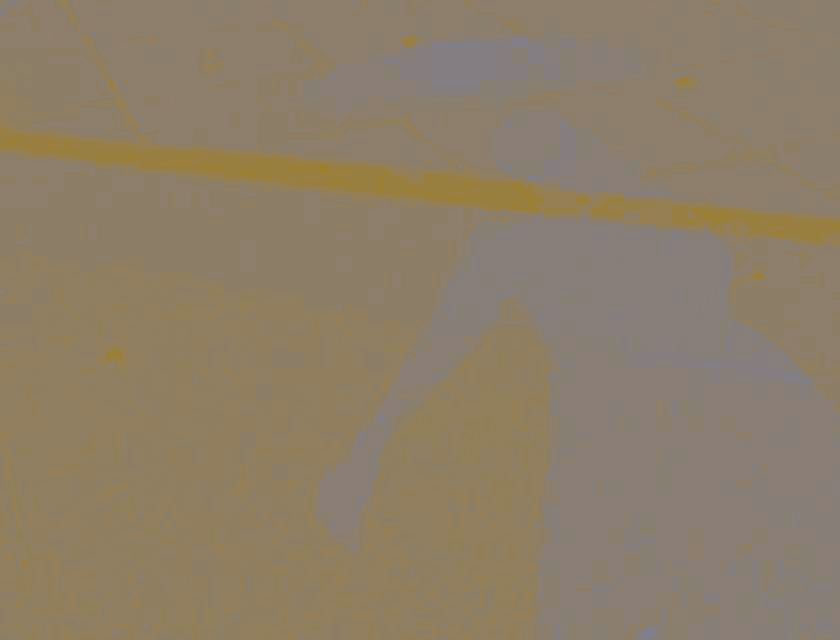} &
        \includegraphics[width=0.093\textwidth]{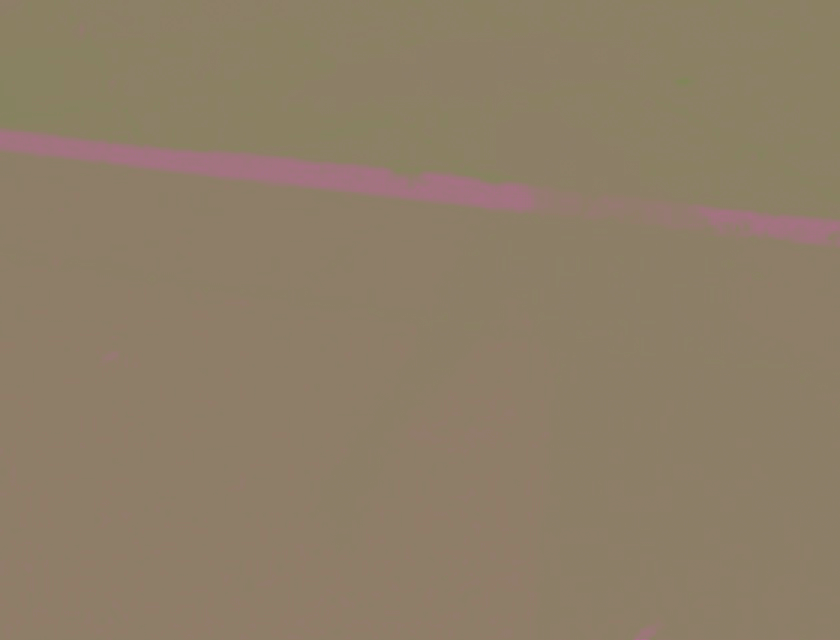} &
        \includegraphics[width=0.093\textwidth]{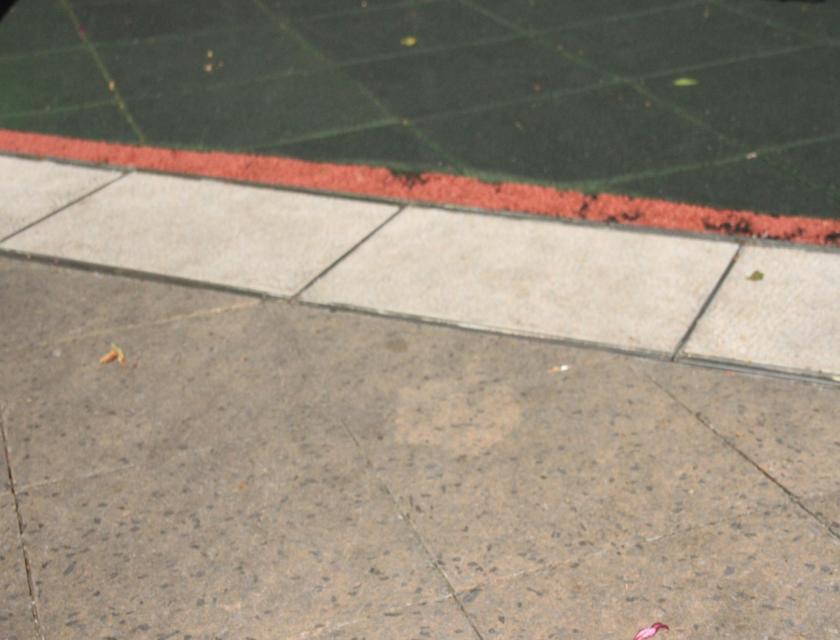} &
        \includegraphics[width=0.093\textwidth]{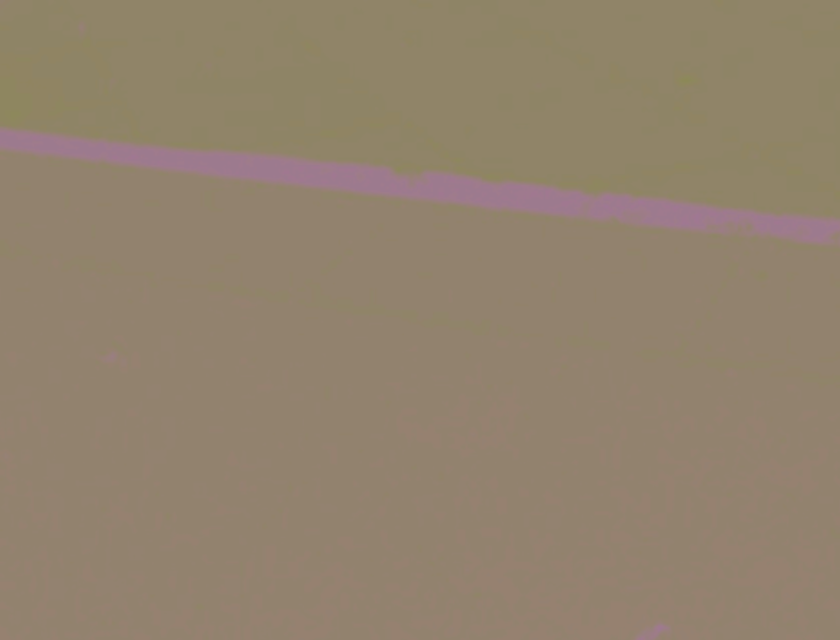} & \\
        \parbox{0.093\textwidth}{\centering \small (a) Input} &
        \parbox{0.093\textwidth}{\centering \small (b) \( \sigma_{s}^{phy} \)} &
        \parbox{0.093\textwidth}{\centering \small (c) \( \sigma_{s}^{IC} \)} &
        \parbox{0.093\textwidth}{\centering \small (d) Output} &
        \parbox{0.093\textwidth}{\centering \small (e) \( \sigma_{sf}^{O} \)} & \\
    \end{tabular}
    \caption{
        (a) Input shadow image, 
        (b) based on shadow-free chromaticity loss \( \sigma_{s}^{\text{phy}} \), 
        (c) Shadow brightness-chromaticity loss (ours) \( \sigma_{sf}^{\text{IC}} \), 
        (d) output shadow-free image, and 
        (e) chromaticity map of the output image \( \sigma_{sf}^{O} \). 
        Compared with shadow-free chromaticity loss, our shadow brightness-chromaticity loss can make the shadow map closer to the shadow-free map, thus helping to remove shadows better.
    }
    \label{fig:network_pipeline} 
\end{figure}

\subsection{Shadow Mask Align Loss}
To ensure that images generated in masked regions have consistent statistical properties with the target image, we introduce a loss function \(\mathcal{L}_{\text{Align}}\). 
\begin{table*}[htbp]
\caption{Quantitative results on AISTD: PSNR, RMSE, SSIM, and LPIPS metrics for entire image (All), shadow regions (S), and shadow-free regions (NS). "$\uparrow$" / "$\downarrow$" indicates that higher/lower is better. Best results are highlighted in \textbf{bold}.}
\label{label1}
\resizebox{\linewidth}{!}{
\begin{tabular}{|l|l|lll|lll|llll|}
\hline
\multicolumn{1}{|c|}{} &
  \multicolumn{1}{c|}{} &
  \multicolumn{3}{c|}{Shadow Region (S)} &
  \multicolumn{3}{c|}{Shadow-Free Region (NS)} &
  \multicolumn{4}{c|}{All Image} \\ \cline{3-12}
\multicolumn{1}{|c|}{Learning} &
  \multicolumn{1}{c|}{Method} &
  PSNR$\uparrow$ &
  RMSE$\downarrow$ &
  SSIM$\uparrow$ &
  PSNR$\uparrow$ &
  RMSE$\downarrow$ &
  SSIM$\uparrow$ &
  PSNR$\uparrow$ &
  RMSE$\downarrow$ &
  SSIM$\uparrow$ &
  LPIPS$\downarrow$ \\ \hline
\multirow{3}{*}{Supervised} & 
  G2R-ShadowNet \cite{liu2021g2rshadow} &
  26.24 &
  15.31 &
  0.962 &
  32.46 &
  3.43 &
  0.946 &
  22.58 &
  5.30 &
  0.876 &
  0.140 \\
 & 
  Auto\cite{guo2012paired} &
  31.00 &
  9.44 &
  0.971 &
  29.32 &
  4.37 &
  0.841 &
  24.14 &
  5.17 &
  0.768 &
  0.174 \\
 &
  Param+M+D-Net \cite{le2020shadow} &
  30.99 &
  10.50 &
  \textbf{0.985} &
  34.50 &
  3.74 &
  0.976 &
  26.58 &
  4.81 &
  \textbf{0.942} &
  0.062 \\ \hline
\multirow{5}{*}{Unsupervised} & 
  Mask-ShadowGAN \cite{hu2019mask} &
  29.37 &
  12.50 &
  0.901 &
  31.65 &
  4.00 &
  0.943 &
  24.57 &
  5.30 &
  0.915 &
  0.095 \\
 & 
  S3R-Net \cite{kubiak2024s3r} &
  - &
  12.16 &
  - &
  - &
  6.38 &
  - &
  - &
  7.12 &
  - &
  - \\
 & 
  LG-ShadowNet \cite{liu2021shadow} &
  30.32 &
  10.35 &
  0.982 &
  32.53 &
  4.03 &
  0.973 &
  25.53 &
  5.03 &
  0.928 &
  0.103 \\
 & 
  DC-ShadowNet \cite{jin2021dc} &
  31.06 &
  10.30 &
  0.978 &
  27.03 &
  3.50 &
  0.971 &
  25.03 &
  4.60 &
  0.921 &
  0.170 \\
 & 
  \textbf{FASR-Net(ours)} &
  \textbf{31.89} &
  \textbf{8.61} &
  0.982 &
  \textbf{34.57} &
  \textbf{2.84} &
  \textbf{0.978} &
  \textbf{27.58} &
  \textbf{3.75} &
  0.934 &
  \textbf{0.055} \\ \hline
\end{tabular}
}
\end{table*}

\textbf{Soft Mask Computation:}
To generate the soft mask, we normalize the shadow and shadow-free images to [0, 1], compute their difference, apply a threshold at the 5th percentile, and set lower pixels to zero. The difference map is then normalized to [-1, 1] to get \( M_1 \), and the soft mask \( M \) is created as \( M = [M_1, M_1, M_1] \).

The soft mask is crucial for mask reconstruction loss, as it aligns the generated soft mask with the ground truth using perceptual and weighted mean square error losses. The perceptual loss, based on VGG-16 features~\cite{johnson2016perceptual}, is given by \( \mathcal{L}_{\text{perceptual}} = \frac{1}{N} \sum_{i=1}^{N} (F_p^i - F_t^i)^2 \).

For the weighted mean square error loss \cite{mathieu2015deep}, we assign a weight of two to pixels in \( M_p \) where \( M_p^{ij} > 0.5 \) to emphasize potential shadow regions, and one otherwise. Thus, the loss formula is \( \mathcal{L}_{MSE-w} = \frac{1}{N} \sum_{i=1}^{N} W_{ij} (M_p^{ij} - M_t^{ij})^2 \), where \( W_{ij} \) doubles the loss for likely shadow pixels.

The smoothness loss, which reduces sharp changes in the mask for smoother transitions, is defined as \( \mathcal{L}_{smooth} = \frac{1}{N} \sum_{i=1}^{N} \left| \nabla (M \odot S_f) \right|_1 \), where \( \nabla \) is the gradient operator, \( \odot \) is element-wise multiplication, and \( \left| \cdot \right|_1 \) is the L1 norm.

Overall, the loss function we optimize for image mask generation is:
\begin{equation}
\mathcal{L}_{\text{recon}} = \mathcal{L}_{\text{smooth}} + \mathcal{L}_{\text{perceptual}} + \mathcal{L}_{\text{MSE-w}} + 0.01 \times \|M\|_2^2.
\end{equation}
\vspace{-1em}

\textbf{Mask Align Loss:}
Using a binary mask, we segment the image into masked and unmasked regions and calculate their mean and standard deviation. Adjusting these statistics for the masked region, we obtain the adjusted (\(\mu_{s}^{\text{os}}\)) and true (\(\mu_{s}^{\text{ws}}\)) masked regions. The alignment loss is computed using the log-cosine loss~\cite{wu2012strong} function:
\begin{equation}
\mathcal{L}_{\text{Align}} = \frac{1}{k} \sum_{l=1}^{k} \ln \left( \cosh \left( \mu_{s}^{\text{os}} - \mu_{s}^{\text{ws}} \right) \right).
\end{equation}

\textbf{Overall Loss} To obtain the overall loss function, each loss function is multiplied by its respective weight and then summed together. The weights corresponding to the losses \(\{\mathcal{L}_{\text{frequency}}, \mathcal{L}_{\text{bright-ch}}, \mathcal{L}_{\text{align}}, \mathcal{L}_{\text{domcls}}, \mathcal{L}_{\text{adv}}, \mathcal{L}_{\text{cons}}, \mathcal{L}_{\text{iden}}\}\) are denoted by \(\{\lambda_{\text{frequency}}, \lambda_{\text{bright-ch}}, \lambda_{\text{align}}, \lambda_{\text{dom}}, \lambda_{\text{adv}}, \lambda_{\text{cons}}, \lambda_{\text{iden}}\}\).

\section{EXPERIMENTS}

\subsection{Experimental Setup}
\textbf {Datasets:} We conduct experiments on two datasets as most prior works: AISTD~\cite{le2019shadow} and SRD~\cite{qu2017deshadownet}. The ISTD dataset, commonly used for supervised shadow removal, contains 1330 training and 540 testing triplets. The AISTD dataset resolves illumination issues between shadow and shadow-free images in the original ISTD dataset. The SRD dataset has 2,680 training pairs and 408 testing pairs of shadow and corresponding shadow-free images. It has no manually annotated masks. 

\textbf {Evaluation Metrics:} For quantitative evaluation, we use RMSE for SRD and AISTD datasets and PSNR for ablation experiments. For SRD, a threshold segmentation method (threshold = 30) is used to obtain the shadow mask, and then the RMSE of shadowed, shadow-free areas and the whole image is measured. For AISTD, the RMSE of these areas is calculated based on the dataset-provided shadow masks.

\textbf {Baselines:} Our proposed method is compared with several existing models. For the AISTD dataset, we compare our method with S3R-Net~\cite{kubiak2024s3r}, LG-ShadowNet~\cite{liu2021shadow}, DC-ShadowNet~\cite{jin2021dc}, and Mask-ShadowGAN~\cite{hu2019mask}. Additionally, we compare with supervised learning methods such as G2R-ShadowNet~\cite{liu2021g2rshadow}, Param+M+D-Net~\cite{le2020shadow}, and Auto~\cite{guo2012paired}. For the SRD dataset, we compare our method with Inpaint4Shadow~\cite{li2023leveraging}, BMNet~\cite{cui2022bmnet}, DSC~\cite{hu2018direction}, G2R-ShadowNet~\cite{liu2021g2rshadow}, Mask-ShadowGAN~\cite{hu2019mask}, DC-ShadowNet~\cite{jin2021dc}.

\textbf{Implementation and Training:} Our FASR-Net is trained unsupervised on a single GPU 4090 for 550,000 iterations with a mini-batch size of 1. Images are randomly cropped to 256×256 from original 480×640, 3-channel images. Training uses a learning rate of 0.0003 and weight decay of 0.0001. Loss weights are set to $\lambda_{\text{brightness-ch}} = 1.1$, $\lambda_{\text{frequency}} = 0.3$, and $\lambda_{\text{Align}} = 0.01$. The network features a base channel number of 64, 4 resblocks, and 6 discriminator layers.


\begin{figure*}[ht]
    \centering
    \setlength{\tabcolsep}{1pt}
    \renewcommand{\arraystretch}{0} 

    \begin{tabular}{cccccccc}
        \includegraphics[width=0.121\textwidth]{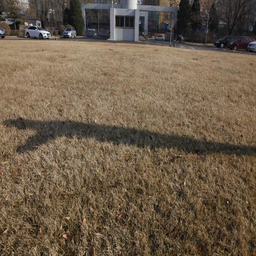} &
        \includegraphics[width=0.121\textwidth]{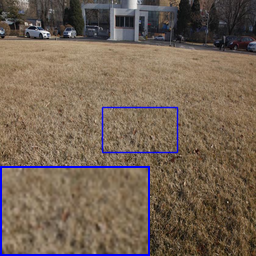} &
        \includegraphics[width=0.121\textwidth]{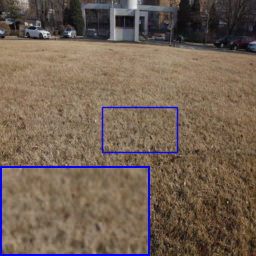} &
        \includegraphics[width=0.121\textwidth]{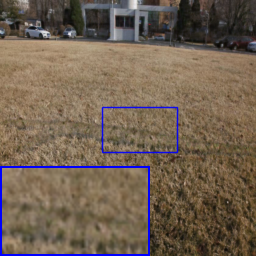} &
        \includegraphics[width=0.121\textwidth]{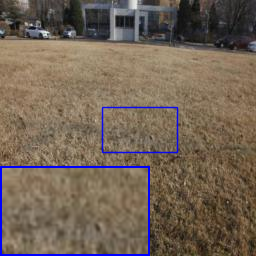} &
        \includegraphics[width=0.121\textwidth]{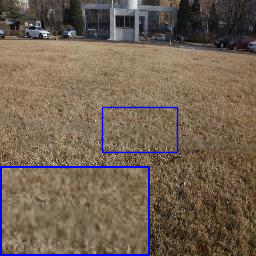} &
        \includegraphics[width=0.121\textwidth]{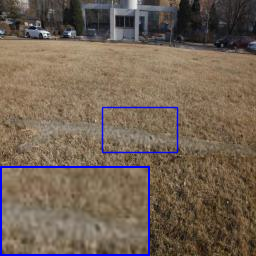} &
        \includegraphics[width=0.121\textwidth]{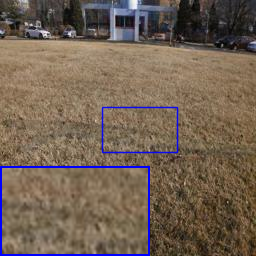} \\
    \end{tabular}

    \vspace{0.2\baselineskip} 

    \vspace{0.2\baselineskip} 

    \begin{tabular}{cccccccc}
        \includegraphics[width=0.121\textwidth]{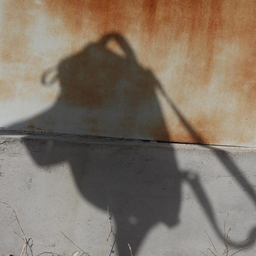} &
        \includegraphics[width=0.121\textwidth]{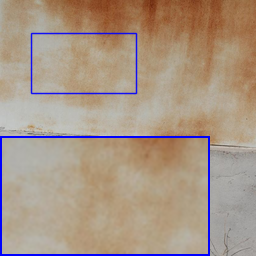} &
        \includegraphics[width=0.121\textwidth]{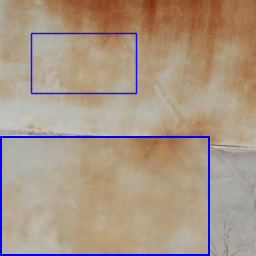} &
        \includegraphics[width=0.121\textwidth]{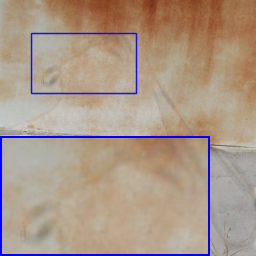} &
        \includegraphics[width=0.121\textwidth]{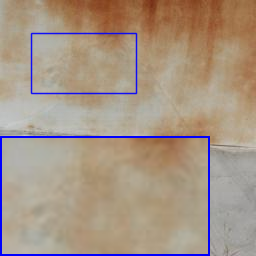} &
        \includegraphics[width=0.121\textwidth]{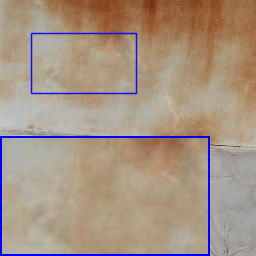} &
        \includegraphics[width=0.121\textwidth]{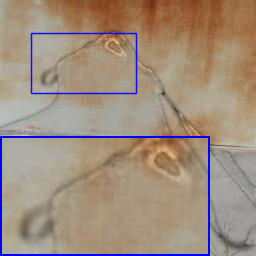} &
        \includegraphics[width=0.121\textwidth]{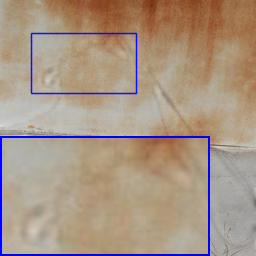} \\
        \parbox{0.121\textwidth}{\centering \small \vspace{1mm}(a) Input} &
        \parbox{0.121\textwidth}{\centering \small \vspace{1mm}(b) Groundtruth} &
        \parbox{0.121\textwidth}{\centering \small \vspace{1mm}(c) Ours} &
        \parbox{0.121\textwidth}{\centering \small \vspace{1mm}(d) DC} &
        \parbox{0.121\textwidth}{\centering \small \vspace{1mm}(e) ST-CGAN} &
        \parbox{0.121\textwidth}{\centering \small \vspace{1mm}(f) DSC} &
        \parbox{0.121\textwidth}{\centering \small \vspace{1mm}(g) G2R} &
        \parbox{0.121\textwidth}{\centering \small \vspace{1mm}(h) Mask-GAN} \\
    \end{tabular}
    \caption{\textbf{Shadow removal results.} Comparison results on the soft shadow SRD dataset. (a) Input image, (b) Groundtruth, (c) Our result, (d) DC-shadowNet, (e) ST-CGAN, (f) DSC, (g) G2R-ShadowNet, and (h) Mask-ShadowGAN. Our unsupervised learning approach produces superior shadow-free results.}
\end{figure*}

\subsection{Experimental Analysis}
As shown in Table \ref{label1}, our FASR-Net, trained without supervision, achieved superior performance on the AISTD dataset compared to existing unsupervised and early supervised baseline methods. Specifically, in the aspect of shadow region removal, relative to DC-ShadowNet~\cite{jin2021dc}, our model attains a 2.7\% increment in PSNR and a 1.69 decrement in RMSE. In addition, FASR-Net maintains excellent capabilities in the restoration process of shadow-free regions and overall images. We also provide visual comparison results in Fig. 5. DC-shadowNet~\cite{jin2021dc} can not restore the details of the shadow-free environment well, as shown in Fig. 5 (d). This is because the information of the image itself is ignored when processing the input image, resulting in incomplete shadow removal. For traditional supervised methods, G2R-ShadowNet~\cite{liu2021g2rshadow} cannot completely remove shadows due to the lack of chromaticity information. The color of the resulting shadow part does not match the original image, as shown in Fig. 5 (g). In contrast, our designed WADM and related loss functions can better solve the problem of inconsistent chromaticity illumination with the surrounding environment. It achieves more precise detail restoration, resulting in shadow removal outcomes.

In the case of the SRD dataset, we expanded our comparison to include recent supervised shadow removal methods. As highlighted in Table \ref{label2}, FASR-Net performed well among unsupervised techniques and outperformed some supervised methods. For example, FASR-Net achieved superior outcomes in the shadow, shadow-free regions, and across the entire image compared to DSC~\cite{hu2018direction}. However, since FASR-Net operates without precise mask inputs for shadow region restoration, its RMSE scores were lower than the most advanced supervised methods of the same period. Although it excels in unsupervised contexts, its dependence on approximate masks limits its performance compared to leading supervised methods like Inpaint4Shadow\cite{li2023leveraging}, which utilize sophisticated mask techniques.\looseness=-1

\begin{table}[htbp]
\centering
\caption{RMSE results on SRD: All, S, and NS denote entire, shadow, and shadow-free regions.}
\label{label2}
\begin{tabular}{llccc}
\toprule
Learning & Method & S & NS & All \\
\midrule
\multirow{4}{*}{SL}
& G2R-ShadowNet  & 11.78 & 4.84 & 6.64 \\
& DSC  & 8.62 & 4.41 & 5.71 \\
& BMNet  & 6.61 & 3.61 & 4.46 \\
& \textbf{Inpaint4Shadow}  & \textbf{6.09} & \textbf{2.97} & \textbf{3.83} \\
\midrule
\multirow{3}{*}{UL} & Mask-ShadowGAN  & 11.46 & 4.29 & 6.40\\
& DC-ShadowNet  & 7.73 & 3.60 & 4.77\\
& \textbf{FASR-Net(ours)} & \textbf{7.45} & \textbf{3.49} & \textbf{4.62}\\
\bottomrule
\end{tabular}
\end{table}

\subsection{Ablation Study}
We perform ablation studies to evaluate the impact of various components within our approach, including the brightness-chromaticity loss \(\mathcal{L}_{brightness-ch}\), shadow frequency loss \(\mathcal{L}_{frequency}\), and shadow mask align loss \(\mathcal{L}_{Align}\). Utilizing the AISTD dataset, we present the quantitative outcomes in Table \ref{label3}. Each of these components plays a significant role in enhancing the performance of our method.

\begin{table}[htbp]
\centering
\small 
\renewcommand{\arraystretch}{0.8} 
\caption{Ablation experiments of our method on SRD}
\begin{tabular*}{\columnwidth}{@{\extracolsep{\fill}}lcccc}
\toprule
Method & \multicolumn{1}{c}{PSNR $\uparrow$} & \multicolumn{1}{c}{RMSE $\downarrow$} & \multicolumn{1}{c}{SSIM $\uparrow$} & \multicolumn{1}{c}{LPIPS $\downarrow$} \\
\midrule
FASR-Net & \textbf{31.89} & \textbf{8.61} & \textbf{0.973} & \textbf{0.033} \\
w/o \(\mathcal{L}_{Align}\) & 30.82 & 9.72 & 0.973 & 0.035 \\
w/o \(\mathcal{L}_{brightness-ch}\) & 30.21 & 10.33 & 0.910 & 0.037 \\
w/o \(\mathcal{L}_{frequency}\) & 29.74 & 10.42 & 0.971 & 0.093 \\
\bottomrule
\end{tabular*}
\label{label3}
\end{table}

\section{Conclusion}
In this paper, we present FASR-Net, a novel unsupervised shadow removal network. Our method employs the Wavelet Attention Downsampling Module (WADM) to leverage frequency prior knowledge, along with a shadow frequency loss, to capture intrinsic shadow characteristics. The brightness-chromaticity loss in the LAB color space enhances color accuracy, while the shadow mask alignment loss ensures feature coherence between shadowed and shadow-free regions. Results on the AISTD and SRD datasets show that our network without mask input exhibited exceptional performance in key shadow removal metrics, such as PSNR and RMSE. It outperformed previous unsupervised methods and surpassed some earlier supervised methods. Despite these successes, future work may explore deeper shadow priors and integrate precise physical models for further enhancement.

\section*{ACKNOWLEDGMENT}

This work is funded in part by the Yunnan Fundamental Research Projects under Grant 202401AW070019, in part by the Youth Project of the National Natural Science Foundation of China under Grant 62201237, in part by the Major Science and Technology Projects in Yunnan Province under Grant 202302AG050009. (Corresponding author: Qingwang Wang.)

\bibliographystyle{IEEEbib}
\bibliography{root} 

\begin{thebibliography}{10}

\bibitem{guo2024single}
Laniqng Guo, Chong Wang, Yufei Wang, Yi~Yu, Siyu Huang, Wenhan Yang, Alex~C Kot, and Bihan Wen,
\newblock ``Single-image shadow removal using deep learning: A comprehensive survey,''
\newblock {\em arXiv preprint arXiv:2407.08865}, 2024.

\bibitem{liu2021g2rshadow}
Zhihao Liu, Hui Yin, Xinyi Wu, Zhenyao Wu, Yang Mi, and Song Wang,
\newblock ``From shadow generation to shadow removal,''
\newblock in {\em CVPR}, 2021, pp. 4927--4936.

\bibitem{liu2023decoupled}
Jiawei Liu, Qiang Wang, Huijie Fan, Wentao Li, Liangqiong Qu, and Yandong Tang,
\newblock ``A decoupled multi-task network for shadow removal,''
\newblock {\em IEEE Transactions on Multimedia}, vol. 25, pp. 9449--9463, 2023.

\bibitem{hu2018direction}
Xiaowei Hu, Lei Zhu, Chi-Wing Fu, Jing Qin, and Pheng-Ann Heng,
\newblock ``Direction-aware spatial context features for shadow detection,''
\newblock in {\em CVPR}, 2018, pp. 7454--7462.

\bibitem{duan2022shadow}
Puhong Duan, Shangsong Hu, Xudong Kang, and Shutao Li,
\newblock ``Shadow removal of hyperspectral remote sensing images with multiexposure fusion,''
\newblock {\em IEEE Transactions on Geoscience and Remote Sensing}, vol. 60, pp. 1--11, 2022.

\bibitem{koutsiou2024sushe}
Dimitra-Christina~C Koutsiou, Michalis~A Savelonas, and Dimitris~K Iakovidis,
\newblock ``Sushe: simple unsupervised shadow removal,''
\newblock {\em Multimedia Tools and Applications}, vol. 83, no. 7, pp. 19517--19539, 2024.

\bibitem{dong2024shadowrefiner}
Wei Dong, Han Zhou, Yuqiong Tian, Jingke Sun, Xiaohong Liu, Guangtao Zhai, and Jun Chen,
\newblock ``Shadowrefiner: Towards mask-free shadow removal via fast fourier transformer,''
\newblock in {\em Proceedings of the IEEE/CVF Conference on Computer Vision and Pattern Recognition}, 2024, pp. 6208--6217.

\bibitem{zeng2024semantic}
Ziqi Zeng, Chen Zhao, Weiling Cai, and Chenyu Dong,
\newblock ``Semantic-guided adversarial diffusion model for self-supervised shadow removal,''
\newblock {\em arXiv preprint arXiv:2407.01104}, 2024.

\bibitem{hu2019mask}
Xiaowei Hu, Yitong Jiang, Chi-Wing Fu, and Pheng-Ann Heng,
\newblock ``Mask-shadowgan: Learning to remove shadows from unpaired data,''
\newblock in {\em ICCV}, 2019, pp. 2472--2481.

\bibitem{vasluianu2021shadow}
Florin-Alexandru Vasluianu, Andr{\'e}s Romero, Luc Van~Gool, and Radu Timofte,
\newblock ``Shadow removal with paired and unpaired learning,''
\newblock in {\em CVPR}, 2021, pp. 826--835.

\bibitem{jin2021dc}
Yeying Jin, Aashish Sharma, and Robby~T Tan,
\newblock ``Dc-shadownet: Single-image hard and soft shadow removal using unsupervised domain-classifier guided network,''
\newblock in {\em ICCV}, 2021, pp. 5027--5036.

\bibitem{yu2023fsr}
Jun Yu, Peng He, and Ziqi Peng,
\newblock ``Fsr-net: Deep fourier network for shadow removal,''
\newblock in {\em Proceedings of the 31st ACM MM}, 2023, pp. 2335--2343.

\bibitem{xu2023haar}
Guoping Xu, Wentao Liao, Xuan Zhang, Chang Li, Xinwei He, and Xinglong Wu,
\newblock ``Haar wavelet downsampling: A simple but effective downsampling module for semantic segmentation,''
\newblock {\em Pattern Recognition}, vol. 143, pp. 109819, 2023.

\bibitem{porwik2004haar}
Piotr Porwik and Agnieszka Lisowska,
\newblock ``The haar-wavelet transform in digital image processing: its status and achievements,''
\newblock {\em Machine graphics and vision}, vol. 13, no. 1/2, pp. 79--98, 2004.

\bibitem{xia2022vision}
Zhuofan Xia, Xuran Pan, Shiji Song, Li~Erran Li, and Gao Huang,
\newblock ``Vision transformer with deformable attention,''
\newblock in {\em CVPR}, 2022, pp. 4794--4803.

\bibitem{misra2021rotate}
Diganta Misra, Trikay Nalamada, Ajay~Uppili Arasanipalai, and Qibin Hou,
\newblock ``Rotate to attend: Convolutional triplet attention module,''
\newblock in {\em WACV}, 2021, pp. 3139--3148.

\bibitem{jiang2021focal}
Liming Jiang, Bo~Dai, Wayne Wu, and Chen~Change Loy,
\newblock ``Focal frequency loss for image reconstruction and synthesis,''
\newblock in {\em ICCV}, 2021, pp. 13919--13929.

\bibitem{horadam2012hadamard}
Kathy~J Horadam,
\newblock {\em Hadamard matrices and their applications},
\newblock Princeton university press, 2012.

\bibitem{chen2021canet}
Zipei Chen, Chengjiang Long, Ling Zhang, and Chunxia Xiao,
\newblock ``Canet: A context-aware network for shadow removal,''
\newblock in {\em ICCV}, 2021, pp. 4743--4752.

\bibitem{mackiewicz1993principal}
Andrzej Ma{\'c}kiewicz and Waldemar Ratajczak,
\newblock ``Principal components analysis (pca),''
\newblock {\em Computers \& Geosciences}, vol. 19, no. 3, pp. 303--342, 1993.

\bibitem{finlayson2009entropy}
Graham~D Finlayson, Mark~S Drew, and Cheng Lu,
\newblock ``Entropy minimization for shadow removal,''
\newblock {\em IJCV}, vol. 85, no. 1, pp. 35--57, 2009.

\bibitem{guo2012paired}
Ruiqi Guo, Qieyun Dai, and Derek Hoiem,
\newblock ``Paired regions for shadow detection and removal,''
\newblock {\em IEEE transactions on pattern analysis and machine intelligence}, vol. 35, no. 12, pp. 2956--2967, 2012.

\bibitem{le2020shadow}
Hieu Le and Dimitris Samaras,
\newblock ``From shadow segmentation to shadow removal,''
\newblock in {\em Computer Vision--ECCV 2020: 16th European Conference, Glasgow, UK, August 23--28, 2020, Proceedings, Part XI 16}. Springer, 2020, pp. 264--281.

\bibitem{kubiak2024s3r}
Nikolina Kubiak, Armin Mustafa, Graeme Phillipson, Stephen Jolly, and Simon Hadfield,
\newblock ``S3r-net: A single-stage approach to self-supervised shadow removal,''
\newblock in {\em CVPRW}, 2024, pp. 5898--5908.

\bibitem{liu2021shadow}
Zhihao Liu, Hui Yin, Yang Mi, Mengyang Pu, and Song Wang,
\newblock ``Shadow removal by a lightness-guided network with training on unpaired data,''
\newblock {\em IEEE Transactions on Image Processing}, vol. 30, pp. 1853--1865, 2021.

\bibitem{johnson2016perceptual}
Justin Johnson, Alexandre Alahi, and Li~Fei-Fei,
\newblock ``Perceptual losses for real-time style transfer and super-resolution,''
\newblock in {\em ECCV}. Springer, 2016, pp. 694--711.

\bibitem{mathieu2015deep}
Michael Mathieu, Camille Couprie, and Yann LeCun,
\newblock ``Deep multi-scale video prediction beyond mean square error,''
\newblock {\em arXiv preprint arXiv:1511.05440}, 2015.

\bibitem{wu2012strong}
Qi~Wu, Wende Zhang, and BVK~Vijaya Kumar,
\newblock ``Strong shadow removal via patch-based shadow edge detection,''
\newblock in {\em ICRA}. IEEE, 2012, pp. 2177--2182.

\bibitem{le2019shadow}
Hieu Le and Dimitris Samaras,
\newblock ``Shadow removal via shadow image decomposition,''
\newblock in {\em ICCV}, 2019, pp. 8578--8587.

\bibitem{qu2017deshadownet}
Liangqiong Qu, Jiandong Tian, Shengfeng He, Yandong Tang, and Rynson~WH Lau,
\newblock ``Deshadownet: A multi-context embedding deep network for shadow removal,''
\newblock in {\em CVPR}, 2017, pp. 4067--4075.

\bibitem{li2023leveraging}
Xiaoguang Li, Qing Guo, Rabab Abdelfattah, Di~Lin, Wei Feng, Ivor Tsang, and Song Wang,
\newblock ``Leveraging inpainting for single-image shadow removal,''
\newblock in {\em ICCV}, 2023, pp. 13055--13064.

\bibitem{cui2022bmnet}
Wenju Cui, Caiying Yan, Zhuangzhi Yan, Yunsong Peng, Yilin Leng, Chenlu Liu, Shuangqing Chen, Xi~Jiang, Jian Zheng, and Xiaodong Yang,
\newblock ``Bmnet: A new region-based metric learning method for early alzheimer’s disease identification with fdg-pet images,''
\newblock {\em Frontiers in Neuroscience}, vol. 16, pp. 831533, 2022.

\end{thebibliography}

\end{document}